\documentclass{article}

\usepackage[final]{nips_2017}

\usepackage[utf8]{inputenc} % allow utf-8 input
\usepackage[T1]{fontenc}    % use 8-bit T1 fonts
\usepackage{hyperref}       % hyperlinks
\usepackage{url}            % simple URL typesetting
\usepackage{booktabs}       % professional-quality tables
\usepackage{amsfonts}       % blackboard math symbols
\usepackage{nicefrac}       % compact symbols for 1/2, etc.
\usepackage{microtype}      % microtypography
\usepackage{graphicx}
\usepackage{amsmath}
\usepackage[export]{adjustbox}
\usepackage{times}
\usepackage[round]{natbib}
\usepackage{multirow}
\DeclareMathOperator*{\argmaxA}{arg\,max}

\title{Learning Rare Word Representations\\ using Semantic Bridging}

\author{
  Victor Prokhorov \\
  Department of Computer Science\\
  University of Cambridge\\
  \texttt{vp361@cam.ac.uk} \\
   \And
  Mohammad Taher Pilehvar  \\
   DTAL\thanks{Department of Theoretical and Applied Linguistics}\\
  University of Cambridge\\
  \texttt{mp792@cam.ac.uk} \\   
  \And
  Dimitri Kartsaklis \\
  DTAL$^*$\\
  University of Cambridge\\
  \texttt{dk426@cam.ac.uk} \\
   \And
   Pietro Li\'{o} \\
  Department of Computer Science\\
  University of Cambridge\\
  \texttt{pl219@cam.ac.uk} \\
   \And
  Nigel Collier \\
  DTAL$^*$\\
  University of Cambridge\\
  \texttt{nhc30@cam.ac.uk} \\
}

\begin{document}

\maketitle

\begin{abstract}
  We propose a methodology that adapts graph embedding techniques (DeepWalk \citep{deepwalk:2014} and node2vec \citep{node2vec:2016}) as well as cross-lingual vector space mapping approaches (Least Squares and Canonical Correlation Analysis) in order to merge the corpus and ontological sources of lexical knowledge. We also  perform comparative  analysis of the  used algorithms in order to identify the best combination for the proposed system.
We then apply this to the task of enhancing the coverage of an existing word embedding's vocabulary with rare and unseen words.
We show that our technique can provide considerable extra coverage (over 99\%), leading to consistent performance gain (around 10\% absolute gain is achieved with \textsc{w2v-gn-500K} cf.\S \ref{ssec:rws}) on the Rare Word Similarity dataset.
\end{abstract}

\section{Introduction}

The prominent model for representing semantics of words is the distributional vector space model \citep{TurneyPantel:2010} and the prevalent approach for constructing these models is the distributional one which assumes that semantics of a word can be predicted from its context, hence placing words with similar contexts in close proximity to each other in an imaginary high-dimensional vector space.
Distributional techniques, either in their conventional form which compute co-occurrence matrices \citep{TurneyPantel:2010,BaroniLenci:2010} and learn high-dimensional vectors for words, or the recent neural-based paradigm which directly learns latent low-dimensional vectors, usually referred to as embeddings \citep{Lecun2015deep}, rely on a multitude of occurrences for each individual word to enable accurate representations.
%KBs are difficult to construct and update, as it requires human expertise and time. 
As a result of this statistical nature, words that are infrequent or unseen during training, such as domain-specific words, will not have reliable embeddings.
This is the case even if massive corpora are used for training, such as the 100B-word Google News dataset \citep{Mikolovetal:2013}.

Recent work on embedding induction has mainly focused on morphologically complex rare words and has tried to address the problem by learning transformations that can transfer a word's semantic information to its morphological variations, hence inducing embeddings for complex forms by breaking them into their sub-word units \citep{luong-socher-manning:2013,Botha2014,soricut-och:2015}.
However, these techniques are unable to effectively model single-morpheme words for which no sub-word information is available in the training data, essentially ignoring most of the rare domain-specific entities which are crucial in the performance of NLP systems when applied to those domains.

On the other hand, distributional techniques generally ignore all the lexical knowledge that is encoded in dictionaries, ontologies, or other lexical resources.
There exist hundreds of high coverage or domain-specific lexical resources which contain valuable information for infrequent words, particularly in domains such as health.
Here, we present a methodology that merges the two worlds by benefiting from both expert-based lexical knowledge encoded in external resources as well as statistical information derived from large corpora, enabling vocabulary expansion not only for morphological variations but also for infrequent single-morpheme words.
The contributions of this work are twofold:
(1) we propose a technique that induces embeddings for rare and unseen words by exploiting the information encoded for them in an external lexical resource, and (2) we apply, possibly for the first time, vector space mapping techniques, which are widely used in multilingual settings, to map two lexical semantic spaces with different properties in the same language.
We show that a transfer methodology can lead to consistent improvements on a standard rare word similarity dataset.

\section{Methodology}
\label{sec:method}

We take an existing semantic space $\mathcal{S}_\mathcal{C}$ and enrich it with rare and unseen words on the basis of the knowledge encoded for them in an external knowledge base (KB) $\mathcal{K}$.
The procedure has two main steps:
we first embed $\mathcal{K}$ to transform it from a graph representation into a vector space representation (\S \ref{sec:kb_embedding}), and then map this space to $\mathcal{S}_\mathcal{C}$ (\S \ref{sec:transformaiton}). Our methodology is illustrated in Figure 1.

In our experiments, we used WordNet 3.0 \citep{Fellbaum:98} as our external knowledge base $\mathcal{K}$.
For word embeddings, we experimented with two popular models: (1) \textsc{GloVe} embeddings trained by \cite{Penningtonetal:2014} on Wikipedia and Gigaword 5 (vocab: 400K, dim: 300), and (2) \textsc{w2v-gn}, Word2vec \citep{Mikolovetal:2013} trained on the Google News dataset (vocab: 3M, dim: 300).
\begin{figure}[h!]

\includegraphics[width=14cm,height=5cm, center]{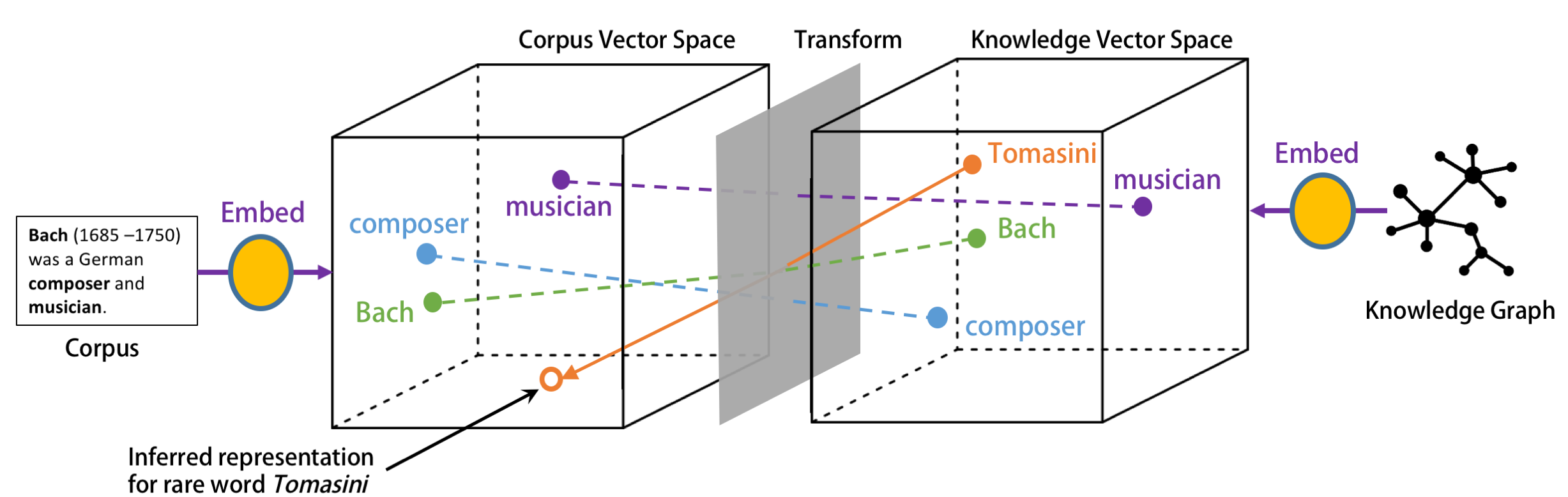}

    \caption{Illustration of the embedding coverage enhancement system. The dashed lines represent \emph{semantic bridges} and the solid line represents a rare word that is transformed from a knowledge vector space to a corpus vector space (\S \ref{sec:transformaiton}).}
    \label{fig:comparison}
\end{figure}

\subsection{Knowledge Base Embedding}
\label{sec:kb_embedding}

Our coverage enhancement starts by transforming the knowledge base $\mathcal{K}$ into a vector space representation that is comparable to that of the corpus-based space $\mathcal{S}_\mathcal{C}$.
To this end, we use two techniques for learning low-dimensional feature spaces from knowledge graphs: \textbf{DeepWalk}  and \textbf{node2vec}.
DeepWalk\footnote{\url{https://github.com/phanein/deepwalk}} uses a stream of short random walks in order to extract local information for a node from the graph. 
By treating these walks as short sentences and phrases in a special language, the approach learns latent representations for each node.
Similarly, node2vec\footnote{\url{https://github.com/snap-stanford/snap/tree/master/examples/node2vec}} learns a mapping of nodes to continuous vectors that maximizes the likelihood of preserving network neighborhoods of nodes.
Thanks to a flexible objective that is not tied to a particular sampling strategy, node2vec reports improvements over DeepWalk on multiple classification and link prediction datasets.
For both these systems we used the default parameters and set the dimensionality of output representation to 100.
Also, note than nodes in the semantic graph of WordNet represent synsets. Hence, a polysemous word would correspond to multiple nodes.
In our experiments, we use the \textit{MaxSim} assumption of \cite{ReisingerMooney:2010} in order to map words to synsets.

To verify the reliability of these vector representations, we carried out an experiment on three standard word similarity datasets: RG-65 \citep{RG65:1965}, WordSim-353 similarity subset \citep{Agirreetal:2009}, and SimLex-999 \citep{Hilletal:2015}. 
Table \ref{table:kb-embedding} reports Pearson and Spearman correlations for the two KB embedding techniques (on WordNet's graph) and, as baseline, for our two word embeddings, i.e. \textsc{w2v-gn} and \textsc{GloVe}.
The results are very similar, with node2vec proving to be slightly superior. 
We note that the performances are close to those of state-of-the-art WordNet approaches \citep{PilehvarNavigli:2015aij}, which shows the efficacy of these embedding techniques in capturing the semantic properties of WordNet's graph.

\begin{table}[t!]
\setlength{\tabcolsep}{21pt}
\begin{center}
\small
\scalebox{0.9}
{
\begin{tabular}{lcccccc}
\toprule
\multirow{1}{*}{\bf KB/Word} & \multicolumn{2}{c}{\bf RG-65} & \multicolumn{2}{c}{\bf WSS-353} & \multicolumn{2}{c}{\bf SimLex-999} \\
\cmidrule(lr){2-3}
\cmidrule(lr){4-5}
\cmidrule(lr){6-7}
\bf Embedding & $\rho$ & $r$ &   $\rho$ & $r$ &   $\rho$ & $r$ \\
\midrule
node2vec    & 0.88      &  0.86  & 0.67  &0.70 & 0.36  &0.39   \\
DeepWalk    & 0.86      &  0.86 & 0.69  &0.70 &0.35   &0.38     \\
\textsc{w2v-gn}   & 0.75       & 0.77      &0.77   &0.76   & 0.44  & 0.45      \\
\textsc{GloVe}       & 0.76      &0.75       &0.66   &0.66   &0.37   & 0.39     \\
\bottomrule

\end{tabular}
}
\end{center}
\caption{\label{table:kb-embedding}Pearson ($r$) and Spearman ($\rho$) correlation results on three word similarity datasets.}
\end{table}

\begin{figure*}[t!]

\includegraphics[trim = 0mm 10mm 0mm 0mm,,width=14.2cm,height=3.8cm, center]{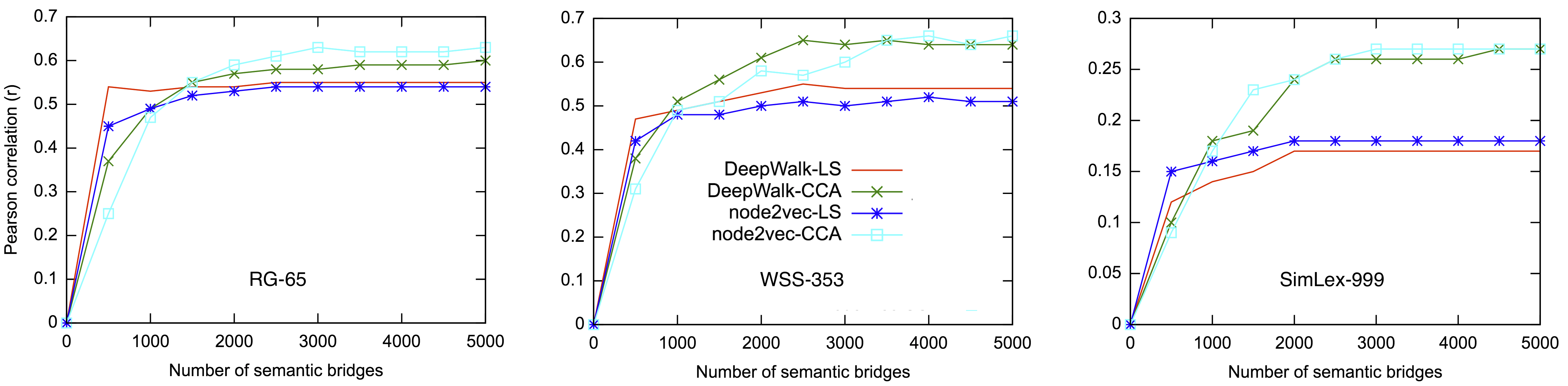}

    \caption{Pearson correlation performance of four configurations: two KB embedding techniques (DeepWalk and node2vec) and two transformation approaches (LS and CCA) on three word similarity datasets.}
    \label{fig:comparison}
\end{figure*}

\subsection{Semantic Space Transformation}
\label{sec:transformaiton}

Once we have the lexical resource $\mathcal{K}$ represented as a vector space $\mathcal{S}_\mathcal{K}$, we proceed with projecting it to $\mathcal{S}_\mathcal{C}$ in order to improve the word coverage of the latter with additional words from the former.
In this procedure we make two assumptions. Firstly, the two spaces provide reliable models of word semantics; hence, the relative within-space distances between words in the two spaces are comparable.
Secondly, there exists a set of shared words between the two spaces, which we refer to as \textit{semantic bridges}, from which we can learn a projection that maps one space into another.

As for the mapping, we used two techniques which are widely used for the mapping of semantic spaces belonging to different languages, mainly with the purpose of learning multilingual semantic spaces: Least squares \cite[\bf LS]{Mikolovetal:13,Dinuetal:2014} and Canonical Correlation Analysis \cite[\bf CCA]{faruqui-dyer:2014,upadhyay-EtAl:2016}.
These models receive as their input two vector spaces of two different languages and a seed lexicon for that language pair and learn a linear mapping between the two spaces.
Ideally, words that are semantically similar across the two languages will be placed in close proximity to each other in the projected space.
We adapt these models to the monolingual setting and for mapping two semantic spaces with different properties.
As for the seed lexicon (to which in our setting we refer to as \textit{semantic bridges}) in this monolingual setting, we use the set of monosemous words in the vocabulary which are deemed to have the most reliable semantic representations.

Specifically, let $\mathcal{S}_\mathcal{C}$ and $\mathcal{S}_\mathcal{K}$ be the corpus and KB semantic spaces, respectively, and  $\mathcal{S}'_\mathcal{C} \subset \mathcal{S}_\mathcal{C}$ and $\mathcal{S}'_\mathcal{K} \subset \mathcal{S}_\mathcal{K}$ be their corresponding subset of \textit{semantic bridges}, i.e., words that are monosemous according to the WordNet sense inventory.
Note that $\mathcal{S}'_\mathcal{C}$ and $\mathcal{S}'_\mathcal{K}$ are vector matrices that contain representations for the same set of corresponding words, i.e., $|\mathcal{S}'_\mathcal{C}| = |\mathcal{S}'_\mathcal{K}|$.
LS views the problem as a multivariate regression and learns a linear function $\mathbf{M} \in \mathbb{R}^{d_{\mathcal{K}} \times d_{\mathcal{C}}}$ (where $d_{\mathcal{K}}$ and $d_{\mathcal{C}}$ are the dimensionalities of the KB and corpus spaces, respectively) on the basis of the following $L_2$-regularized least squares error objective and typically using stochastic gradient descent:
\begin{equation}
    \min_{\mathbf{M} \in \mathbb{R}^{d_{\mathcal{K}} \times d_{\mathcal{C}}}} ||\mathcal{S}'_\mathcal{K}\mathbf{M} - \mathcal{S}'_\mathcal{C}||^2 + \lambda ||\mathbf{M}||^2
\end{equation}

The enriched space $\mathcal{S}^*$ is then obtained as a union of $\mathcal{S}_\mathcal{K}\mathbf{M}$ and $\mathcal{S}_\mathcal{C}$.
CCA, on the other hand, learns two distinct linear mappings $\mathbf{M}_1$ and $\mathbf{M}_2$ with the objective of maximizing the correlation between the dimensions of the projected vectors $\mathbf{M}_1 \mathcal{S}'_\mathcal{C}$ and $\mathbf{M}_2 \mathcal{S}'_\mathcal{K}$:
\begin{equation}
\begin{split}
    \mathbf{M}_1,\mathbf{M}_2 &= CCA(\mathcal{S}'_\mathcal{K},\mathcal{S}'_\mathcal{C}) \\
                              &=  \argmaxA_{\mathbf{M}_1,\mathbf{M}_2} \rho (\mathbf{M}_1 \mathcal{S}'_\mathcal{C}, \mathbf{M}_2 \mathcal{S}'_\mathcal{K})
\end{split}
\end{equation}
In this case, $\mathcal{S}^*$ is the union of $\mathbf{M}_1 \mathcal{S}_\mathcal{C}$ and $\mathbf{M}_2 \mathcal{S}_\mathcal{K}$.
In the next section we first compare different KB embedding and transformation techniques introduced in this section and then apply our methodology to a rare word similarity task.

\section{Experiments}
\subsection{Evaluation benchmark}
To verify the reliability of the transformed semantic space, we propose an evaluation benchmark on the basis of word similarity datasets.
Given an enriched space $\mathcal{S}^*$ and a similarity dataset $D$, we compute the similarity of each word pair $(w_1, w_2) \in D$ as the cosine similarity of their corresponding transformed vectors $s_{w_1}$ and $s_{w_2}$ from the two spaces, 
where $s_{w_1} \in \mathcal{S}_\mathcal{K}\mathbf{M}$ and $s_{w_2} \in \mathcal{S}_\mathcal{C}$ for LS and $s_{w_1} \in \mathbf{M}_1 \mathcal{S}_\mathcal{C}$ and $s_{w_2} \in \mathbf{M}_2 \mathcal{S}_\mathcal{K}$ for CCA.
A high performance on this benchmark shows that the mapping has been successful in placing semantically similar terms near to each other whereas dissimilar terms are relatively far apart in the space.
We repeat the computation for each pair in the reverse direction.
\subsection{Comparison Study}

Figure \ref{fig:comparison} shows the performance of different configurations on our three similarity datasets and for increasing sizes of semantic bridge sets.
We experimented with four different configurations: two KB embedding approaches, i.e. DeepWalk and node2vec, and two mapping techniques, i.e. LS and CCA (cf. \S \ref{sec:method}).
In general, the optimal performance is reached when around 3K semantic bridges are used for transformation.
DeepWalk and node2vec prove to be very similar in their performance across the three datasets.
Among the two transformation techniques, CCA consistently outperforms LS on all three datasets when provided with 1000 or more semantic bridges (with 500, however, LS always has an edge).
In the remaining experiments we only report results for the best configuration: node2vec with CCA.
We also set the size of semantic bridge set to 5K.\footnote{We used only monosemous nouns and adjectives as our semantic bridges (WordNet 3.0 has over 100K of these). Our experiments with sets upto 20K semantic bridges did not show any significant performance gain over 5K.}

\subsection{Rare Word Similarity}
\label{ssec:rws}
In order to verify the reliability of our technique in coverage expansion for infrequent words we did a set of experiments on the Rare Word similarity dataset \citep{luong-socher-manning:2013}.
The dataset comprises 2034 pairs of rare words, such as \textit{ulcerate}-\textit{change} and \textit{nurturance}-\textit{care}, judged by 10 raters on a [0,10] scale.
Table \ref{table:rw-sim} shows the results on the dataset for three pre-trained word embeddings (cf. \S \ref{sec:method}), in their initial form as well as when enriched with additional words from WordNet.\footnote{In addition to our two pre-trained embeddings, we also experimented with the top 500K words from \textsc{w2v-gn} in order to simulate a setting with limited vocabulary.}

Among the three initial embeddings, \textsc{w2v-gn-500K} provides the lowest coverage, with over 20\% out-of-vocabulary pairs, whereas \textsc{GloVe} has a similar coverage to that of \textsc{w2v-gn} despite its significantly smaller vocabulary (400K vs. 3M).
Upon enrichment, all the embeddings attain near full coverage (over 99\%), thanks to the vocabulary expansion by rare words in WordNet.
The enhanced coverage leads to consistent performance improvements according to both Pearson and Spearman correlations.
The best performance gain is achieved for \textsc{w2v-gn-500K} (around 10\% absolute gain) which proves the efficacy of our approach in inducing embeddings for rare words. The improvements are also statistically significant (p < 0.05) according to conducted one tailed t-test \citep{Coh:75}, showing that the coverage enhancement could lead to improved performance even if lower-performing KB embedding and transformation are used.

\begin{table}[t!]
\setlength{\tabcolsep}{17pt}
\begin{center}
\small
\scalebox{0.97}
{
\begin{tabular}{lcccccc}
\toprule
\bf  \multirow{2}{*}{Word embedding} & \multicolumn{3}{c}{\bf Initial} &
\multicolumn{3}{c}{\bf Enriched} \\
\cmidrule(lr){2-4}
\cmidrule(lr){5-7}
&  OOV &  $r$ &   $\rho$ & OOV & $r$ &   $\rho$  \\
\midrule
\textsc{w2v-gn}   & 173
	&	{0.44}	&0.45	&~9	&\bf 0.46	&\bf 0.48	\\
\midrule
\textsc{w2v-gn-500K}  & 453
	&	0.36	& 0.34		&16	&\bf  0.42	&\bf 0.44	\\
\midrule
\textsc{GloVe}	&  216      
	&   0.35 & 0.34	&~6	&\bf 0.37	&\bf	0.38\\
\bottomrule

\end{tabular}
}
\end{center}
\caption{\label{table:rw-sim}Results on the Rare Word Similarity dataset for different word embeddings, before and after enrichment.}
\end{table}

\section{Related Work}

The main focus of research in embedding coverage enhancement has been on the morphologically complex forms \citep{alexandrescu-kirchhoff:2006}.
\cite{luong-socher-manning:2013} used recursive neural networks (RNNs) and neural language
models in order to induce embeddings for morphologically complex words from their morphemes whereas \cite{lazaridouetal:2013} adapted phrase composition models for this purpose.
\cite{Botha2014} proposed a different model based on log-bilinear language models, mainly to have a compositional vector-based morphological representation that can be easily integrated into a machine translation decoder.
These models often utilize a morphological segmentation toolkit, such as Morfessor \citep{CreutzLagus:2007}, in order to break inflected words into their morphological structures and to obtain segmentations for words in the vocabulary.
\cite{soricut-och:2015} put forward a technique that 
does not rely on any external morphological
analyzer and instead, induces
morphological rules and transformations, represented
as vectors in the same embedding space.
Based on these rules a morphological graph is constructed and representations are inferred by analyzing morphological transformations in the graph.
However, all these techniques fall short in inducing representations for single-morpheme words that are not seen frequently during training as they base their modeling on information available on sub-word units.
In contrast, our transformation-based model can also induce embeddings for single-morpheme words that are infrequent or unseen in the training data, such as domain-specific entities.

\section{Conclusions and Future Work}

We presented a methodology for merging distributional semantic spaces and lexical ontologies and applied it to the task of extending the vocabulary of the former with the help of information extracted from the latter.
We carried out an analysis for different KB embedding and semantic space mapping techniques and also showed that our methodology can lead to considerable enrichment of two standard word embedding models, leading to consistent improvements on the rare word similarity dataset.
One interesting property of our approach is that it can be used in the reverse direction and for the completion of knowledge bases using the distributional information derived from text corpora.
In future work, we plan to investigate this direction.
We also intend to experiment with domain-specific lexical resources and measure the impact of coverage enhancement on a downstream NLP application.

\subsubsection*{Acknowledgments}
This research was supported by EPSRC Experienced Researcher Fellowship (Nigel Collier, Dimitri Kartsaklis, (EP/M005089/1)), MRC grant (Mohammad Taher Pilehvar, (MR/M025160/1))
We gratefully acknowledge the donation of a GPU card from the NVIDIA
Grant Program.

%\section*{References}
%\bibliographystyle{unsrtnat}

%\bibliography{rareWords}
\end{document}